\def\year{2019}\relax
\documentclass[letterpaper]{article} 
\usepackage{aaai19}  
\usepackage{times}  
\usepackage{helvet} 
\usepackage{courier}  
\usepackage[hyphens]{url}  
\usepackage{graphicx} 
\usepackage{graphicx}  
\usepackage{xcolor}
\usepackage{soul}
\usepackage[utf8]{inputenc}
\usepackage{float}

\usepackage{url}  
\usepackage{amsmath}
\usepackage{comment}
\usepackage{float}
\usepackage{amsfonts}
\usepackage{booktabs}
\usepackage{siunitx}
\usepackage[linesnumbered,ruled]{algorithm2e}

\usepackage{amssymb}
\usepackage{times}

\def\reals{{\mathbb R}}
\newcommand{\nnreals}{\reals_{\ge 0}}
\newcommand\mafs{$\sf MAFS$}
\newcommand\mafsb{$\sf MAFSB$}
\newcommand\masiw{$\sf MA$-$\sf SIW$}
\newcommand\iw{$\sf IW$}
\newcommand\siw{$\sf SIW$}
\newcommand\bfws{$\sf BFWS$}
\newcommand\mabfws{$\sf MA$-$\sf BFWS$}
\newcommand\kmabfws{$\sf k$-$\sf MA$-$\sf BFWS$}
\newcommand\hffp{$h_{\sf FF}^P$}
\newcommand\hff{$h_{\sf FF}$}

\newcommand\maplan{$\sf MAPLAN$}
\newcommand\psm{$\sf PSM$}
\newcommand\dpp{$\sf DPP$}

\newcommand\blocksworld{$\rm Blocksworld$}
\newcommand\depot{$\rm Depot$}
\newcommand\driverlog{$\rm DriverLog$}
\newcommand\elevators{$\rm Elevators$} 
\newcommand\logistics{$\rm Logistics$} 
\newcommand\rovers{$\rm Rovers$}
\newcommand\satellites{$\rm Satellites$}
\newcommand\sokoban{$\rm Sokoban$}
\newcommand\taxi{$\rm Taxi$}
\newcommand\wireless{$\rm Wireless$}
\newcommand\woodworking{$\rm Woodworking$} 
\newcommand\zenotravel{$\rm Zenotravel$}
\newcommand\mabw{$\rm MA$-$\rm Blocksworld$}
\newcommand\mabwhard{$\rm MA$-$\rm Blocksworld$-$\rm Large$}
\newcommand\malog{$\rm MA$-$\rm Logistics$}
\newcommand\maloghard{$\rm MA$-$\rm Logistics$-$\rm Large$}
\newcommand\mabwshort{$\rm MA$-$\rm BW$}
\newcommand\mabwhardshort{$\rm MA$-$\rm BW$-$\rm L$}
\newcommand\malogshort{$\rm MA$-$\rm Log$}
\newcommand\maloghardshort{$\rm MA$-$\rm Log$-$\rm L$}

\usepackage[linesnumbered,ruled]{algorithm2e}
\makeatletter
\newenvironment{Balgorithm}[1][htpb]
  {\def\@algocf@post@ruled{\kern\interspacealgoruled\hrule  height\algoheightrule\kern3pt\relax}%
    \def\@algocf@capt@ruled{under}
    \begin{algorithm}[#1]}
  {\end{algorithm}}
\makeatother
\newtheorem{definition}{Definition}
\newtheorem{theorem}{Theorem}
\newtheorem{lemma}{Lemma}
\newtheorem{corollary}{Corollary}

\newcommand{\tup}[1]{\langle #1\rangle}            
\SetKwProg{Fn}{Procedure}{}{}

\title{Best-First Width Search for Multi Agent Privacy-preserving Planning}
\author{Alfonso E. Gerevini \\ Dipartimento di Ingegneria dell'Informazione \\ Universit\`a degli Studi di Brescia, Italy
        \And Nir Lipovetzky \\ School of Computing and Information Systems \\ The University of Melbourne\\
        \AND Francesco Percassi \and Alessandro Saetti \and Ivan Serina\thanks{Corresponding author. Email: ivan.serina@unibs.it}  \\ Dipartimento di Ingegneria dell'Informazione \\ Universit\`a degli Studi di Brescia, Italy
}

\newif\ifLONGPAPER
\LONGPAPERfalse 

\begin{document}

\maketitle
\begin{abstract}
In multi-agent planning, preserving the agents' privacy has become an increasingly popular research  topic. For preserving the agents' privacy, agents jointly compute a plan that achieves mutual goals by keeping certain information private to the individual agents. Unfortunately, this can severely restrict the accuracy of the heuristic functions used while searching for solutions. 
It has been recently shown that, for centralized planning, the performance of goal oriented search can be improved by combining goal oriented search and width-based search. The combination of these techniques has been called best-first width search. 
In this paper, we investigate the usage of best-first width search in the context of (decentralised) multi-agent privacy-preserving planning, addressing the challenges related to the agents' privacy and performance. In particular, we show that best-first width search is a very effective approach over several benchmark domains, even when the search is driven by heuristics that roughly estimate the distance from goal states, computed without using the private information of other agents. An experimental study analyses the effectiveness of our techniques and compares them with the state-of-the-art.
\end{abstract}

\section{Introduction}
Over the last years, several frameworks for multi-agent (MA) planning have been proposed, e.g., \cite{brafman2008one,nissim2014distributed,torreno2014fmap}.  Most of them consider, in different ways, the agents' privacy: some or all agents have private knowledge that cannot be communicated to other agents during the planning process and  plan execution. This prevents the straightforward usage of most of the current powerful techniques developed for centralized (classical) planning, which are based on heuristic functions computed by using the knowledge of all the involved agents.

For classical planning, it has been shown that width-based search algorithms can solve instances of many existing domains in low polynomial time when they feature atomic goals. Width-based search relies on the notion of ``novelty''. The novelty of a state $s$ has been originally defined as the size of the smallest tuple of facts that holds in $s$ for the first time in the search, considering all previously generated states \cite{LipovetzkyG12}. 
Width-based search are pure exploration methods that are not goal oriented. For computing plans that are not necessarily optimal, the performance of goal oriented search can be improved by combining it with width-based search. The combination yields a search procedure, called best-first width search, that outperforms the state-of-the-art planners even when the estimate of the distance to the problem goals is inaccurate \cite{DBLP:conf/aaai/LipovetzkyG17}. The heuristic used to guide such a procedure uses {\em all} the knowledge of the problem specification. 

In the setting of MA planning, computing search heuristics using the knowledge of all the involved agents can require many exchanges of information among agents, and this may compromise the agents' privacy. For preserving the privacy of the involved agents, the distance to the problem goals is estimated by using the knowledge of a {\em single} agent. However, this estimate is much more inaccurate than the estimation obtainable using the knowledge of all the agents. Since for classical planning best-first width search performs very well even when the estimate of the goal distance is inaccurate, such a procedure is a good candidate to effectively solve MA-planning problems without compromising the agents' privacy.
The contribution of the paper is investigating the usage of best-first width search for decentralized privacy-preserving MA-planning. Specifically, we propose a new search procedure \mabfws, which uses width-based exploration in the form of novelty-based preferences to provide a complement to goal-directed heuristic search. 

For preserving the privacy of the involved agents, the private knowledge shared among agents is encrypted. An agent $\alpha_i$ shares with the other agents a description of its search states in which all the private facts of $\alpha_i$ that are true in a state are substituted with a string code obtained by encrypting all the private fact names of $\alpha_i$ together. This encryption has an impact on the measure of novelty, and hence it also affects the definition of the heuristic guiding the search. 

We adapt the definition of classical width~\cite{LipovetzkyG12} to MA planning, and we propose a definition of state novelty for which \mabfws\ can be complete when states are pruned only if their novelty is bigger than the width of the problem. Then, we define several heuristics for which the preferred states in the open list are the ones with the smallest novelty and, among those, the ones with the lowest goal distance. For this purpose, we define the novelty in another different way, taking the heuristics used to estimate the goal distance into account \cite{DBLP:conf/aaai/LipovetzkyG17}. 

Finally, an experimental study evaluates the effectiveness of the proposed heuristics, and compares the proposed approach with state-of-the-art planners which preserve agents' privacy in a form weaker than our approach, showing that best-first width search is competitive also for MA planning.

\section{Background}\label{sec:background}

\vspace{1mm}
\paragraph{The MA-STRIPS planning problem.} Our work relies on {\it MA-STRIPS}, a ``minimalistic'' extension of the STRIPS language for MA planning \cite{brafman2008one}, that is the basis of the most popular definition of MA-planning problem (see, e.g., \cite{nissim2014distributed,maliah2016stronger}). 

\begin{definition}
A  {\sc MA-STRIPS} {planning problem} $\Pi$ for a set of agents $\Sigma = \{\alpha_i\}_{i = 1}^n$ is a 4-tuple $\langle \{A_i\}_{i=1}^n,~ P,~I,~G  \rangle$ where:

\begin{itemize}
\item $A_i$ is the set of actions agent $\alpha_i$ is capable of executing, and s.t.\
for every pair of agents $\alpha_i$ and $\alpha_j$ $A_i \cap A_j = \emptyset$;   
\item $P$ is a finite set of propositions;
\item $I \subseteq P$ is the initial state;
\item $G \subseteq P$ is the set of goals.

\end{itemize}
\end{definition}

\noindent 
Each action $a$ consists of a name, a set of preconditions, $Prec(a)$, representing facts required to be true for the execution of the action, a set of additive effects, $Add(a)$, representing facts that the action makes true, a set of deleting effects, $Del(a)$, representing facts that the action makes false, and a real number, $Cost(a)$, representing the cost of the action.
A fact is {\it private} for an agent if other agents can neither
achieve, destroy or require the fact \cite{brafman2008one}. A fact is {\it public}
otherwise.   An action is private if all its preconditions and
effects are private; the action is public, otherwise. A state obtained by executing a public action is said to be public; otherwise, it is private.

To maintain agents' privacy, the private knowledge shared among agents can be encrypted. An agent can share with the other agents a description of its search state in which each private fact that is true in a state is substituted with a string obtained by encrypting the fact name \cite{Bonisoli2018}. This encryption of states does not reveal the names of the private facts of each agent $\alpha_i$ to other agents, but an agent can realize the existence of a private fact of agent $\alpha_i$ and monitor its truth value during search. This allows the other agents to infer the existence of private actions of $\alpha_i$, as well as to infer their causal effects. Another way to share states containing private knowledge during the search is to substitute, for each agent $\alpha_i$, all the private facts of $\alpha_i$ that are true in a state with a string obtained by encrypting all private fact names of $\alpha_i$ together \cite{nissim2014distributed}. Such a string denotes a dummy private fact of $\alpha_i$, which is treated by other agents as a regular fact. The work presented in this paper uses this latter method for the encryption of states. With this method,  other agents can only infer the existence of a group of private facts of $\alpha_i$, since the encrypted string contained in the states exchanged by $\alpha_i$ substitutes a group of an arbitrary number of private facts of $\alpha_i$. 

\paragraph{Width-based Search.} Pure width-based search algorithms are exploration algorithms that do
not look at the goal at all. The simplest such algorithm is $IW(1)$,
which is a plain breadth-first search where newly generated states
that do not make an atom $X = x$ true for the first time in the search
are pruned. The algorithm $IW(2)$ is similar except that a state $s$
is pruned when there are no atoms $X = x$ and $Y = y$ such that the
pair of atoms $X = x$, $Y = y$ is true in $s$ and false in all the
states generated before $s$. 

$IW(k)$ is a normal breadth-first except
that newly generated states $s$ are pruned when their ``novelty'' is
greater than $k$, where the novelty of $s$ is $i$ iff there is a tuple
$t$ of $i$ atoms such that $s$ is the first state in the search that
makes all the atoms in $t$ true, with no tuple of smaller size having
this property \cite{LipovetzkyG12}.  While simple, it has been shown
that $IW(k)$ manages to solve arbitrary instances of many of the
standard 
benchmark domains in low polynomial time
provided that the goal is a single atom. Such domains can be shown to
have a small and bounded {\em width} $w$ that does not depend on the
instance size, which implies that they can be solved (optimally) by
running $IW(w)$. Moreover, $IW(k)$ runs in time and space that are
exponential in $k$ and not in the number of problem variables.  

The procedure $IW$, that calls $IW(1)$ and $IW(2)$,
sequentially, has been used to solve instances featuring multiple
(conjunctive) atomic goals, in the context of Serialized IW
(\siw), an algorithm that calls $IW$ for
achieving one atomic goal at a time \cite{LipovetzkyG12}.  

Width-based exploration in the form of simple novelty-based preferences instead of pruning can provide an effective complement to goal-directed heuristic search without sacrificing completeness. Indeed, it has been recently shown that the combination of width-based search and heuristic search, called best-first width search (\bfws), yields a search scheme that is better than both, and outperforms the state-of-the-art planners \cite{DBLP:conf/aaai/LipovetzkyG17}. 

\section{Related Work}

The MA-planning algorithm most similar to ours is \mafs\ \cite{nissim2014distributed}. 
\mafs\ is a distributed best first search that for each agent considers a separate search space.
The existing work investigating the use of a distributed A* for partial-order MA-planning shares the motivations on preserving the agents' privacy with ours  \cite{torreno2014fmap}.
Differently from this approach, our MA-planning procedure searches in the space of world states, rather than in the space of partial plans, and it exchanges states among agents rather than partial plans. 

Our work is also related to the one on developing heuristics for MA-planning. \citeauthor{stolba2015comparison} (\citeyear{stolba2015comparison}) study the use of heuristic functions based on the heuristic of the well-known planner $\sf FF$. Similarly, the work in \cite{stolba2014relaxation} 
proposes a distributed algorithm computing a complete relaxed planning graph and, subsequently, extracting a relaxed plan from the distributed relaxed planning graph. Differently, our heuristics combine the novelty measure of search states with a more inaccurate, but computationally cheaper estimate of the cost required to achieve the problem goals.

Using width-based search for MA planning is not a novel idea. \citeauthor{bazzotti2018iterative} (\citeyear{bazzotti2018iterative}) study the usage of Serialized-IW (abbreviated by \masiw) in the setting of MA planning. The MA problem solved by \masiw\ is split into a sequence of episodes, where each episode $j$ is a subproblem solved by IW, returning a path to a state where one more problem goal has been achieved with respect to the last episode $j-1$. 
Our approach does not split the MA problem into subproblems, but solves the whole problem at once by using the novelty as a heuristic to guide the search. 

\iw\ search was also used for solving a classical planning problem obtained from the compilation of a MA-planning problem \cite{Muise2015}. This work applies to centralized MA planning, while our work investigates the distributed MA-planning problem.

An important difference between our approach and the existing ones using heuristic search for distributed  privacy-preserving MA planning (e.g. \cite{maliah2016stronger,stolba2015admissible}) is that with our approach the public projection of public actions is not shared. Our conjecture is that without sharing such a projection it is more difficult to infer private preconditions and effects of public actions, since the agents ignore their existence. While sharing the public projection of public actions may be useful to compute more accurate search heuristic, our approach is nevertheless competitive with the state-of-the-art planners.

\section{Width-based Search for MA planning}

A problem of Serialized-IW and \masiw\ is that they are incomplete. In this section, we propose another approach of using width-based search for MA planning, which guarantees that a solution is found when the problem is solvable. 

\begin{Balgorithm}[!t]
	
	\SetKwInOut{Input}{Input}
	\SetKwInOut{Output}{Output}
	
	\caption{\kmabfws\ run by agent $\alpha_i$ from the initial state $s_I$ to achieve goals $G$ using only set of actions $A_i$ of  $\alpha_i$. The output is a single-agent solution plan $\pi_i$ for  $\alpha_i$, or failure. Parameter $k \in \mathbb{N}$ is an upper bound for the novelty of expanded states. $g(s)$ is the accumulated cost, and $f$ is the eval function to sort the \textit{open} list.
    } 
	        \Fn{\kmabfws$(s_I,G,A_i,f)$}{

	\label{alg:MA-BFWS}
	{\small
		$open \gets s_I$  
		
        $g_{s_I} \gets 0$

		\While {$open$ \textrm{is not} $ empty$ \textbf{or} $open\_msg$ \textrm{is not} $empty$\footnotemark}{
			
            \ForEach{$s\in open\_msg$}
			{
            $open \gets open \cup s $
            
            $open\_msg \gets open\_msg \setminus s$
			}
			
			$s \gets \textbf{SelectBest}(f,open)$

			$open \gets open \setminus s$

			\If(\tcc*[f]{Plan found}){$G \in s $}{
				$ \textbf{return ReconstructPlan}(s)$
			}
			
			\If{$s$ was generated by agent $\alpha_i$ and $s$ is public}{				
				$ \textbf{SendStateMessage}(\tup{s, g(s)})$
			}			
			\ForEach(\tcc*[f]{\hspace{-1.5mm}Expand\hspace{-1.5mm}}){$a \in A_i$ \textrm{s.t.} $Prec(a) \subseteq s$}
			{
				$s' \gets s \setminus Del(a) \cup Add(a)$

				$g(s') \gets g(s) + Cost(a)$

				\If{$w_{(g)}(s')  \leq k$}
				{
					$open \gets open \cup s'$
				}
			}
		}
		\textbf{return failure}
	}	
	}
\end{Balgorithm}

\footnotetext{More precisely, when the $open$ and $open\_msg$ lists of an agent become empty, the agent sends a special message to the other agents representing the fact that its own lists are empty. Similarly, the agent sends another special message to the others when its own open list is not empty anymore. The algorithm terminates when the lists of all the agents are empty.}

Algorithm \ref{alg:MA-BFWS} shows a search procedure for an agent of the MA-planning problem combining width-based search and goal-directed search, that we call \kmabfws. Parameter $k \in \mathbb{N}$ is an upper bound for the novelty of states that can be expanded, i.e., states with novelty greater than $k$ are pruned from the search space. 
The version of the algorithm without this pruning is called \mabfws; in this version the novelty is used as a preference criterion for ranking the search states in the open list.

Each agent $\alpha_i$ considers a separate search space, since each agent maintains its own list of open states, $open$, and, when an agent expands an open state, it generates a set of successor states using its own actions. Moreover, each agent also maintains its own list of received messages to process, $open\_msgs$. Algorithm \ref{alg:MA-BFWS} assumes the presence of a separate thread listening for incoming messages sent from other agents; each time a message is received, it is added to the end of $open\_msg$ list. 

Agent $\alpha_i$ iteratively expands the states in the open list and those contained in the received messages. Loop 4--23 is repeated until the lists $open$ and $open\_msg$ are  empty. 
Agent $\alpha_i$ extracts all the states in $open\_msg$, computes the novelty according with the states generated or received by $\alpha_i$, computes the given heuristic function $f$, and adds the states to the open list. Then, $\alpha_i$  extracts the best state $s$ from $open$ according to $f$ (steps 5--9). We considered $f$ as a sequence of $n$ arbitrary heuristics $\tup{h_1, \dots, h_n}$ that are applied consecutively to break ties.
Each time a state $s$ is extracted from $open$, first $\alpha_i$ checks if the state satisfies the goals of the planning problem. If it does, agent $\alpha_i$, together with the other agents, reconstructs the plan achieving $s$ and returns its solution single-agent plan (steps 11--13). Once an agent expands a solution state $s$, Procedure ${\small\sf ReconstructPlan}(s)$ performs the trace-back of the solution plan. Agent $\alpha_i$ begins the trace-back, and when it reaches a state received via a message $m$, it sends a trace-back message to the agent who sent $m$. This continues until the initial state is reached. The MA-plan derived from the trace-back is a solution of the MA planning problem. Finally, at step 12 Algorithm \ref{alg:MA-BFWS} returns the plan output by ${\small\sf ReconstructPlan}(s)$.

Then, agent $\alpha_i$ checks if state $s$ is the result of its own public action, and in this case it sends a message to all other agents containing state $s$ together with its accumulated cost $g(s)$ from the initial state up to $s$ (steps 14--16). Finally, $\alpha_i$ expands state $s$ by applying the executable actions and, for each successor state $s'$ of $s$,  $\alpha_i$ evaluates the novelty and evaluation function $f$, and decides whether to add $s'$ in its $open$ list according to the novelty of state $s'$ (steps 17--23). 

Algorithm \ref{alg:MA-BFWS} prunes a state $s$ according to a novelty measure akin to \citeauthor{KatzLMT17} (\citeyear{KatzLMT17}) novelty heuristics, but defined instead on the basis of the cost $g(s)$ accumulated through the trajectory from the problem initial state to $s$ (steps 20-21).

\begin{definition} [Accumulated Cost Novelty]
\label{def:wg}
The novelty $w_{(g)}(s)$ of a state $s$ is the size of the smallest tuple $t$ in $s$ that: (1) is achieved for the first time during search, or (2) for which every other previously generated state $s'$ where $t$ is true has longer paths, i.e., $g(s') > g(s)$. 
\end{definition}

The accumulated cost $g$ of the states that are at the same time in the $open$ list of agent $\alpha_i$ can be very different, because the search does not necessarily extract the state from $open$ with the lowest accumulated cost, and $open$ may contain also states incoming from other agents, who visit different search spaces that might contain states with a much greater $g$-value.  

To guarantee the agents' privacy, the private knowledge contained in the search states exchanged among agents is encrypted. The encryption affects the measure of novelty. E.g., consider states $s_1 = \{p\}$, $s_2=\{q\}$, $s_3 = \{ p, q \}$, where $p$ and $q$ are private facts of an agent different from $\alpha_i$. Let $[x]$ denote the encrypted string representing one or more private facts $x$ of another agent. With the encryption, the descriptions of these states for $\alpha_i$ are $\{[p]\}, \{[q]\}, \{ [p, q] \}$. Assume that the order with which these states are processed by $\alpha_i$ is $s_1$, $s_2$, $s_3$, and $g(s_1) \le g(s_2) \le g(s_3)$. Then, without the encryption, for $\alpha_i$ $w_{(g)}(s_1) = w_{(g)}(s_2) = 1$, $w_{(g)}(s_3) = 2$, while with the encryption we have $w_{(g)}(s_1) = w_{(g)}(s_2) = w_{(g)}(s_3) = 1$, because in $s_3$ the special string representing encrypted facts $[p, q]$ is true for the first time in the search. This consequently affects the pruning of the search space: 1-\mabfws\ without the encryption prunes $s_3$ from the search space, while 1-\mabfws\ encrypting private facts does not prune $s_3$.

\begin{lemma}
\label{noveltylema}
The novelty $w_{(g)}(s)$ computed over the set of previously generated encrypted states is lower than or equal to the novelty computed over the set of previously generated states without the encryption. 
\end{lemma}

\noindent
{\bf Proof.} For simplicity, take a MA-planning problem with only two agents $\alpha_i$ and $\alpha_j$, and consider the computation of the novelty for $\alpha_i$. The proof for problems with more than two agents is similar. Take a state $s$ and a tuple $t \subseteq s$ such that, without the encryption, $w_{(g)}(s) = |t|$. With the encryption, we distinguish three cases. (1) Tuple $t$ is formed by public or private facts of $\alpha_i$. In this case, since only the private facts of $\alpha_j$ are encrypted for $\alpha_i$, the facts forming tuple $t$ are the same as without the encryption, and hence even with the encryption $w_{(g)}(s) = |t|$. (2) Tuple $t$ is formed by at least $n \ge 1$ private facts of agent $\alpha_j$, and the tuple $t'$ of private facts of $\alpha_j$ that are true in $s$ is different from those of previously generated  states such that their accumulated cost is lower than or equal to $g(s)$. Then, with the encryption, the tuple $t'$ is substituted by a new string. Such a string denotes a dummy fact that is false in all the previously generated states. Hence, by definition, with the encryption $w_{(g)}(s) = 1$, and, of course, it is lower than or equal to $|t|$. (3) Tuple $t$ is formed by at least $n \ge 1$ private facts of agent $\alpha_j$, the tuple $t'$ of private facts of $\alpha_j$ that are true in $s$ is the same as in a state $s'$, $s'$ has been previously expanded, and $g(s') \le g(s)$. With the encryption, the tuple $t'$ is substituted by a string $u$, which denotes a dummy fact that, in this case, is true in both $s$ and $s'$. Therefore, the smallest tuple in $s$ that is true for the first time in the search is formed by public or private facts of $\alpha_i$ plus $u$. By definition, with the encryption $w_{(g)}(s) = |t|-n+1$ and, since $n \ge 1$, such a value of $w_{(g)}(s)$ is lower than or equal to $|t|$.  $\Box$

\vspace{2mm}
The definition of width  by \citeauthor{LipovetzkyG12} (\citeyear{LipovetzkyG12}) for the state model induced by STRIPS applies directly to the state model induced by MA-STRIPS. 

\begin{definition}[MA-STRIPS width]
\label{def:width}
The width $w(\Pi)$ of a MA-STRIPS planning problem $\Pi$ is $w$ if there is a sequence of tuples $\langle t_0, \ldots, t_n \rangle$ such that (1) $t_i \subseteq P$ and $|t_i| \leq w$ for $i=0,\ldots,n$,  (2) $t_0 \subseteq I$, (3) all optimal plans for $t_i$ can be extended into optimal plans for $t_{i+1}$ with an action $a\in \{A_i\}_{i=1}^{|\Sigma|}$, and (4) all optimal plans for $t_n$ are also optimal plans for $G$.
\end{definition}

In the definition above, some actions that extend optimal plans for a tuple $t_i$ into optimal plans for tuple $t_{i+1}$ can be private. \kmabfws\ does not send states generated by private actions to other agents. In the following theorem, note that the novelty $w(s)$ of a state $s$ is computed with respect to the search space of  agent $\alpha$ that generated $s$. The search space of $\alpha$ includes its generated states as well as the states received from other agents.

\begin{theorem}
 \kmabfws\ using $f$ = $\langle w, h_1, \ldots, h_m \rangle$ is complete for problems $\Pi$ with width $w(\Pi) \leq k$ when w = $w_{(g)}$ and the action cost function is $Cost : A \mapsto \reals^+$.
\end{theorem} 
\noindent
{\bf Proof Sketch.} 
The definition of width $w(\Pi) = k$ (Def.~\ref{def:width}) implies that there is an optimal plan $\pi^{opt}$ where every state $s_i$ along the plan has novelty $w(s_i) \le k$ (Def.~\ref{def:wg}), inducing a sequence of tuples $\langle t_0, \ldots, t_n \rangle$ that complies with the conditions in Definition~\ref{def:width}. If $w=w_{(g)}$, by Definition~\ref{def:wg}, there must exist at least one tuple $t\in s_i$, where $|t|\le k$, such that no other state $s'$ can be generated with $t\in s'$  and $g(s')<g(s_i)$. \kmabfws\ with $w=w_{(g)}$ is guaranteed to generate each state $s_i$ in $\pi^{opt}$, no matter the order in which states are generated, as long as by assumption zero cost actions are not allowed. State expansion order is determined breaking ties by a sequence of search heuristics $h_i$. No heuristic $h_i$ causes \kmabfws\ to prune nodes, even if $h_i=\infty$, as $h_i$ does not have access to the private actions of other agents and cannot be proved to be \textit{safe}. When a state in an optimal plan has been generated by another agent, the private facts are encoded. Given Lemma~\ref{noveltylema}, the novelty of such states is guaranteed to be lower or equal, hence they are not going to be pruned by \kmabfws.
$\Box$

\begin{theorem}
 \mabfws\ is complete.
\end{theorem}

\noindent
{\bf Proof.} 
\mabfws\ does not prune the search space according to the novelty of search states. In \mabfws, each agent $\alpha_i$ expands all the search states reachable from the problem initial state except the private states of agents different from $\alpha_i$. This is the same set of search states expanded by \mafs. Since \mafs\ is a complete search procedure \cite{nissim2014distributed}, even \mabfws\ is complete. $\Box$ 


\begin{theorem}\label{th:bounds}
	Let $P^{pub}$ be the set of public facts, and $P_i^{pr}$ be the set of private facts of agent $\alpha_i\in\Sigma$ such that the total number of facts $P$ is $P^{pub} \cup_{\alpha_i\in \Sigma}P_i^{pr}$. Let $n_i=\sum_{\alpha_i\in\Sigma, j\neq i} |P_j^{pr}|^k$ be the number of encrypted strings denoting dummy fluents that can be sent to agent $\alpha_i$.  \kmabfws\ using heuristic f terminates after expanding at most
	\begin{enumerate}
		\item $\sum_{\alpha_i\in\Sigma} \big(|P^{pub}| \hspace{-0.2mm}+\hspace{-0.2mm} |P_i^{pr}| \hspace{-0.2mm}+\hspace{-0.2mm} n_i\big)^{k}$ states if action costs are 0,
		\item $\sum_{\alpha_i\in\Sigma} \big(|P^{pub}| \hspace{-0.2mm}+\hspace{-0.2mm} |P_i^{pr}| \hspace{-0.2mm}+\hspace{-0.2mm} n_i\big)^{2k} $ \hspace{-1.4mm}states if action costs are 1, 
		\item $\sum_{\alpha_i\in\Sigma} \big(|P^{pub}| + |P_i^{pr}| + n_i\big)^{2k}  \hspace{-1.4mm}\times |A|$ states when the cost function is $Cost : A \mapsto \nnreals$,
	\end{enumerate} where $|\Sigma|$ is the number of agents, and $|A|$ is the number of available actions for all agents.
\end{theorem} 

\noindent
{\bf Proof.} 
Let $f_i=|P^{pub}| + |P_i^{pr}| + n_i$ be the number of possible state facts for an agent $\alpha_i$ of the MA problem $\Pi$. We distinguish three cases. (1) When all action costs are one, the longest path $\pi$ an agent $\alpha_i$ can expand has length $|\pi| = f_i^k$. For $\pi$ to be expanded, every state $s_1,\ldots,s_{f_i^{k}}$ along the path needs to have novelty $w_g(s_i)\le k$. Therefore, each state $s_i$ either makes a tuple $t$ of size $k$ true for the first time, or achieves a tuple t with a lower $g(s_i)<g(s')$ than other previously generated states $s'$ with $t\in s'$. A path $|\pi| > f_i^{k}$ cannot be expanded as the state $s_{f_i^{k}+1}$ in the path has novelty $w_g(s_{f_i^{k}+1})>k$. For a path to reach length $f_i^k$ , each state $s_1,\ldots,s_{f_i^{k}}$ must have added at most \textit{one} new tuple or improved the $g$-value of at most \textit{one} tuple to pass the novelty pruning criteria  $w_g(s_i)\le k$. Since $g$ grows monotonically, the $g$-value of a tuple $t$ cannot be improved more than once along the same path $\pi$. Once state $s_{f_i^{k}+1}$ is expanded in the path, all tuples have been generated with smaller $g$-values. Therefore, each tuple $t$ can let $f_i^{k}$ states to be expanded with novelty $w_g(s_i)\le k$. Given that we consider at most $f_i^{k}$ tuples, in total we can expand $f_i^{2k}$ states. In the worst case, each agent $\alpha_i \in \Sigma$ can expand the state space independently, yielding the overall $\max_{\alpha_i\in\Sigma}O(f_i^{2k})$. (2) In case all action costs are zero, the $g$-value can never be improved once a tuple has been made true by a state. Therefore each tuple $t$ can let just one state to be expanded with novelty $w_g(s_i)\le k$, and the total number of expanded states is at most $\max_{\alpha_i\in\Sigma}O(f_i^{k})$. (3) If the cost function $Cost : A \mapsto \nnreals$ maps actions $A$ to positive real numbers including zero, then each tuple $t$ can be improved $f_i^{2k}$  with $A_t$ action, the number of actions with different cost that add tuple $t$, which in the worst case is $A$. Therefore the total number of states that can be expanded is $\max_{\alpha_i\in\Sigma}O(f_i^{2k} \times |A|)$.
$\Box$

\begin{table}
\scriptsize
\begin{center}
\begin{tabular}{|l|c|c|c|c|}
\hline 
Domain                         & \#Instances & 1-\mabfws & 2-\mabfws & \hff \\ 
\hline
\blocksworld&214&100.0\%&100.0\%&100.0\%\\ 
\depot&155&85.81\%&91.61\%&100.0\%\\ 
\driverlog&185&95.68\%&100.0\%&100.0\%\\ 
\elevators&255&82.75\%&66.27\%&99.22\%\\ 
\logistics&172&0.0\%&93.6\%&100.0\%\\ 
\rovers&277&98.92\%&100.0\%&99.28\%\\ 
\satellites&488&20.9\%&98.36\%&100.0\%\\ 
\sokoban&61&54.1\%&93.44\%&98.36\%\\ 
\taxi&95&90.53\%&98.95\%&98.95\%\\ 
\wireless&160&51.88\%&36.88\%&58.75\%\\ 
\woodworking&1084&98.89\%&99.17\%&97.05\%\\ 
\zenotravel&258&99.22\%&79.84\%&100.0\%\\ 
\hline 
{\bf Overall} &3404& 77.59\%& 91.63\%& 96.94\%  \\ 
\hline 
\end{tabular}

\end{center}
\caption{\label{tab:pruning}Number of instances, and coverages of 1-\mabfws\ and 2-\mabfws\ guided by $g$ w.r.t.\  \mabfws\ guided by $f=\langle$\hff$\rangle$ computed by each agent using its own actions for problem instances with a single goal.} 
\end{table}

\begin{corollary}
Let $m$ be the maximum novelty of a state expanded by \mabfws, once a plan has been found and \mabfws\  terminates. Then, the number of expanded states is bounded by the complexity of \kmabfws\ when $k=m$.
\end{corollary}

If a problem is solved by 1-\mabfws\ it does not entail that the problem has width 1, it entails rather a lower bound, much like the notion of effective width discussed by \citeauthor{LipovetzkyG12} (\citeyear{LipovetzkyG12}). Still, it provides an estimate on how hard it is to solve a MA planning problem. Table \ref{tab:pruning} shows that, even for the MA setting, for single atom goals all domains but \wireless\ generally have width lower than or equal to 2. For this analysis, we considered the domains from the distributed track of the first international competition on distributed and multi-agent planning. For each instance with $m$ goal atoms, we created $m$ instances with a single goal, and ran 1-\mabfws\ and 2-\mabfws\ over each one of them. The total number of instances is 3404. The search heuristic used for 1-\mabfws\ and 2-\mabfws\ is very simple: the best state in the open list is selected among those with the lowest novelty measure $w_{(g)}$, breaking the ties with the accumulated cost $g$. For each domain we show the total number of single goal instances, and the percentage of instances  solved with width equal to 1 and lower than or equal to 2. We considered action costs unitary. Therefore, by Theorem \ref{th:bounds}, this table shows that 77.59\% of the single goal problems can be solved with a quadratic time $O(n^2)$  where $n=|P|$ is the number of propositions in the problem. These blind and  bounded planners perform well with respect to a baseline goal-directed heuristic search planner, \mabfws\ guided by the same heuristic used by planner $\sf FF$ but computed by each agent using only its own action.
However, problems with multiple goals in general have a higher width. In the next section we explore how to scale up to multiple goals.

\section{Novelty-based heuristics}

In MA planning, agents have private information that they do not want to share with others.  The heuristic computed using only the knowledge of one single agent can be much more inaccurate than using the knowledge of all the agents. 
%
In this section, we propose some search heuristics for privacy-preserving MA planning that combine the measure of the novelty of search states with the estimated distance to reach the problem goals. The goal distance is estimated by using the knowledge of a single agent. Our conjecture is that, in the MA setting, width-based exploration in the form of novelty-based preferences can provide a complement to goal-directed heuristic search, so that the search can be effectively guided towards a goal state even if the goal-directed heuristics are inaccurate. 

The computation and memory cost of determining that the novelty $w$ of a state $s$ is $k$ is exponential in $k$, since all the tuples of size up to $k$ but one may be stored and considered. For efficiency, we simplify the computation of novelty $w(s)$ to only 3 levels, i.e. $w(s)$ is determined to be equal to 1, 2, or greater than 2.  

For our heuristic functions, we used the measure of novelty introduced by \citeauthor{DBLP:conf/aaai/LipovetzkyG17} (\citeyear{DBLP:conf/aaai/LipovetzkyG17}). Given the arbitrary functions $h_1, \dots, h_n$,  the novelty $w(s')$ of a newly generated state $s'$  is $k$ iff there is a tuple of $k$ atoms $X_k = x_k$ and no tuple of smaller size, that is true in $s$ but false in all previously  generated states $s'$ with the same function values $h_1 (s') = h_1 (s), \dots,$ and $h_n (s') = h_n (s)$. For example, a new state $s$ has novelty $1$ if there is an atom $X = x$ that is true in $s$ and false in all the states $s'$ generated before $s$ where $h_i (s') = h_i (s)$ for all $i$. In the rest of the paper, the novelty measure $w$ is sometimes denoted as $w_{(h_1,\dots,h_n)}$  in order to make explicit in the notation the functions $h_i$ used in the definition and computation of $w$. 

The first heuristic we study is $f_1 = \tup{w_{(h_{\sf FF})},h_{\sf FF}}$, where component $h_{\sf FF}$ denotes the goal-directed heuristic used by planner $\sf FF$. The goal distance of an agent $\alpha_i$ from a search state $s$ is estimated as the number of actions of $\alpha_i$ in a relaxed plan constructed from $s$ to achieve the problem goals. The plan is relaxed because it is a solution of a relaxed problem in which the negative action effects are removed. 
Substantially, the best state $s$ in $open$ according to $f_1$ is not selected among those with the lowest estimated goal distance, but among those with the lowest novelty measure $w(s) = w_{(h_{\sf FF})}$, and heuristic \hff\ is only used to break the ties. The same heuristic was proposed for classical planning obtaining, surprisingly, good results \cite{DBLP:conf/aaai/LipovetzkyG17}. The difference with respect to classical planning is that for MA planning the distance estimated to reach the problem goals is much more inaccurate. For an agent $\alpha_i$, the relaxed plan is extracted using only the actions of $\alpha_i$. When an agent evaluates the search states by using only its own set of actions, it is possible that at least one of the goals is evaluated as unreachable. In this case, the extraction of the relaxed plan fails, and the estimated distance is evaluated as infinite. This is due to the agent not being able to solve the problem alone,  needing to cooperate with other agents.

We consider other types of information for the definition of the search heuristic, in order to overcome the problem of the inaccuracy of the goal-directed heuristics computed using only the knowledge of a single agent. In the following, given a search state $s$, $G_\bot$ and $G_u$ denote the number of goals that are false in $s$ and the number of goals that are unreachable from $s$, respectively. For an agent $\alpha_i$, the number of goals unreachable from $s$ is estimated by constructing with the actions of $\alpha_i$ a relaxed planning-graph (RPG) from $s$. The goals that are not contained in the last level of the RPG are considered  unreachable.  

Planner \mabfws\ with heuristic function $f_2 = \tup{w_{(G_\bot, h_{\sf FF})},G_\bot,h_{\sf FF}}$, denoted as \mabfws$(f_2)$, selects the next state $s$ to expand among those in $open$ with the lowest novelty measure  $w(s) = w_{(G_\bot, h_{\sf FF})}$, breaking the ties according to the number of goals that are false in $s$. Heuristic $h_{\sf FF}$ is finally used to break the ties when there is more than one state in $open$ with the same lowest measure of novelty and the same lowest number of false goals.  

Similarly, \mabfws\ with heuristic function $f_3 = \tup{w_{(G_u,G_\bot, h_{\sf FF})},G_u,G_\bot,h_{\sf FF}}$ 
selects the next state $s$ to expand among those in $open$ with the lowest novelty measure  $w(s) = w_{(G_u,G_\bot, h_{\sf FF})}$, breaking  ties according to the number of goals that are unreachable from $s$. If there is more than one state in $open$ with the same lowest measure of novelty and the same lowest number of unreachable goals, the ties are broken according to the number of goals false in $s$. Finally, heuristic $h_{\sf FF}$ is used only if there are still ties to break.

The drawback of $h_{\sf FF}$ for MA planning is that often the estimated goal distance from a search state $s$ is infinite, even though $s$ is not a dead-end. As stated before, the reason for this is that from $s$ the planning problem cannot be solved by an agent alone. With the next search heuristic, we study a method to overcome this problem, for which an agent $\alpha_i$ extracts a relaxed plan from $s$ to the (sub)set of problem goals that are {\em reachable}  from $s$. The estimated distance from $s$ to all the problem goals is the number of actions in the relaxed plan plus the number of problem goals unreachable from $s$ multiplied by a constant. In our experiments, such a constant is equal to the maximum number of levels in the RPGs constructed so far. This variant of $h_{\sf FF}$ is denoted as \hffp. Essentially, the information about the unreachable goals is used to refine the estimated goal distance. We report experiments with  \mabfws$(f_4)$, where the  function $f_4$ is obtained from $f_3$ using \hffp\ in place of $h_{\sf FF}$ as goal-directed component of the evaluation function.


Components $G_u$ and \hffp\ of heuristic $f_4$ are computationally expensive, since for each expanded  state $G_u$ requires the construction of a RPG, and \hffp\ additionally requires the extraction of a relaxed plan from the RPG. The last two heuristics we study consider the tradeoff between the accuracy of the estimated goal-distance and its computational cost. For this, the construction of the RPG and the extraction of the relaxed plan are not performed for each expanded state, but only for the initial state of the planning problem and the search states incoming from other agents. The facts that are preconditions of the actions in the relaxed plan are called {\it relevant}. Let $s'$ be the last incoming state in the way to state $s$ for which the relaxed plan was extracted. For evaluating the goal distance of state $s$, we consider the number $\#r$ of relevant facts that have {\it not} been made true in the way from $s'$ to $s$. This measure is similar to that proposed by \citeauthor{DBLP:conf/aaai/LipovetzkyG17} (\citeyear{DBLP:conf/aaai/LipovetzkyG17}) for classical planning. 
The difference with respect to classical planning is that a relaxed plan is extracted for each incoming state, instead of for the states that decrement the number of achieved problem goals in relation to their parent. This is needed  as the relevant fluents are not sent among agents in order to avoid compromising privacy.
Planner \mabfws$(f_5)$ with $f_5 =  \tup{w_{(G_\bot, \#r)},G_\bot,\#r}$ considers counter $\#r$ in place of the more computationally expensive components $G_u$ and~\hffp.

The drawback of heuristic $f_5$ is that, when the number of exchanged messages is high, it still requires the construction of the RPG many times. The construction of the RPG is computationally much more expensive than extracting the relaxed plan and, when such a construction is performed many times, it can become the bottleneck of the search procedure. Thereby, we propose another heuristic $f_6$ which, for each agent $\alpha_i$, requires the construction of the RPG from only the initial state of the problem. With the aim of maintaining the agents' privacy, the RPG is still constructed by using only the actions of a single agent. Nevertheless, when the MA-planning problem cannot be solved by an agent alone, the last level of the RPG does not contain the problem goals. For this, the construction of the RPG from the initial state is special, and is done in two steps. The first step is the construction of the RPG from $s_I$. Then, in the second step, the preconditions $p\in pre(a)$ of actions $a\in A_i$ of agent $\alpha_i$ that are not added by actions of $\alpha_i$ and are not true in the last level of the RPG, are made true in the last level of the RPG. Finally, the construction of the RPG continues from the last level of the RPG constructed so far. 

Consider a state $s$ to be expanded. For heuristic $f_6$, the counter $\#r$ is defined as the number of relevant facts in the RPG constructed from the problem initial state $s_I$ that have {\it not} been made true in the way from $s_I$ to $s$. The computation of $\#r$ for $f_6$ differs from $f_5$, because an agent $\alpha_i$ alone can reconstruct only the portion of trajectory from the last incoming state $s'$ to $s$, and cannot reconstruct the trajectory from $s_I$ to $s'$. The trajectory from $s_I$ to $s'$ can contain other actions of agent $\alpha_i$ that should be taken into account in the definition of the set of relevant facts that have {\it not} been made true by $\alpha_i$ in the way from $s_I$ to $s$. For this, the presence of these actions of $\alpha_i$ is estimated by solving a {\it super relaxed} planning problem, i.e., a planning problem with the same initial state of the planning problem, the set of facts that are true in $s'$ as goals, and a set of actions obtained from the set of actions of $\alpha_i$ by ignoring the action preconditions that are unreachable from $s_I$, as well as negative action effects. The procedure for the extraction of the super-relaxed plan is similar to the one used by $\sf FF$. The positive effects of the actions in such a super-relaxed plan are facts that we estimate have been made true in the way from $s_I$ to $s'$. Therefore, for $f_6$ we define the counter $\#r$ as the number of relevant facts that have {\it not} been made true in the super-relaxed plan from $s_I$ to $s'$ and in the way from $s'$ to $s$.

 \begin{table}
\scriptsize
\begin{center}
\begin{tabular}{|l|c|c|c|c|c|c|c|}
\hline
Domain& \hff&
$f_1$&
$f_2$ &
$f_{3}$&
$f_{4}$&
$f_{5}$&
$f_{6}$ \\ \hline 
{\bf From CoDMAP}&&&&&&&\\
	\blocksworld&19&{\bf 20}&{\bf 20}&{\bf 20}&{\bf 20}&{\bf 20}&{\bf 20} \\ 
	\depot&5&6&19&19&{\bf 20}&{\bf 20}&{\bf 20} \\ 
	\driverlog&18&{\bf 20}&{\bf 20}&{\bf 20}&{\bf 20}&{\bf 20}&{\bf 20} \\ 
	\elevators&3&3&{\bf 20}&{\bf 20}&{\bf 20}&{\bf 20}&{\bf 20} \\ 
	\logistics&3&3&{\bf 20}&{\bf 20}&{\bf 20}&{\bf 20}&{\bf 20} \\ 
	\rovers&14&17&{\bf 20}&{\bf 20}&{\bf 20}&{\bf 20}&{\bf 20} \\ 
	\satellites&18&19&{\bf 20}&{\bf 20}&{\bf 20}&{\bf 20}&{\bf 20} \\ 
	\sokoban&{\bf 18}&{\bf 18}&{\bf 18}&{\bf 18}&{\bf 18}&{\bf 18}&14 \\ 
	\taxi&{\bf 20}&19&19&{\bf 20}&{\bf 20}&{\bf 20}&{\bf 20} \\ 
	\wireless&{\bf 2}&{\bf 2}&{\bf 2}&{\bf 2}&{\bf 2}&{\bf 2}&{\bf 2} \\ 
	\woodworking&3&3&10&{\bf 15}&{\bf 15}&11&12 \\ 
	\zenotravel&{\bf 20}&{\bf 20}&{\bf 20}&{\bf 20}&{\bf 20}&{\bf 20}&{\bf 20} \\ 
	\hline 
{\bf From MBS}&&&&&&&\\
	\mabwshort&0&0&11&11&12&2&{\bf 19} \\ 
	\malogshort&0&0&0&0&1&0&{\bf 15} \\ 
	\mabwhardshort&0&0&{\bf 20}&{\bf 20}&{\bf 20}&{\bf 20}&19 \\ 
	\maloghardshort&0&0&{\bf 20}&{\bf 20}&{\bf 20}&{\bf 20}&19 \\ 
	\hline 
\bf{Overall} (320)&143&150&259&265&268&253&{\bf 280} \\ 
	\hline 
\end{tabular}
\end{center}
\caption{\label{tab:coverage}Number of problems solved by \mabfws\ with seven different heuristics for the benchmarks problems of CoDMAP and MBS.  The best performance is in bold.}
\end{table}

\section{Experiments}

In this section, we present an experimental study aimed at testing the effectiveness of the heuristics described so far. First, we describe the experimental settings; then we evaluate the effectiveness of our heuristics; finally, we compare the performance of our approach with the state of the art.

Our code is written in C++, and exploits the Nanomsg open-source library to share messages \cite{nanomsg}. 
 Each agent uses three threads, two of which send and receive messages, while the other one conducts the search, so that the search is asynchronous w.r.t.\ the communication routines. The behavior of \mabfws\ depends on the order with which the messages are received by an agent. Each time a run of \mabfws\ is repeated, the agents' threads can be scheduled by the operating system differently, so that the behavior of \mabfws\ can also be different. Thereby, for each problem of our benchmark, we run \mabfws\ five times and consider the performance of the algorithm as the median over the five runs. When \mabfws\ exceeded the CPU-time limit for more than two of the five runs, we consider the problem unsolved.

The benchmark used in our experiments includes twelve domains proposed by  \citeauthor{vstolba2015competition} (\citeyear{vstolba2015competition}) for the distributed track of the first international competition on distributed and MA planning (CoDMAP), and four domains \mabw\ (shortly, \mabwshort), \mabwhard\ (\mabwhardshort), \malog\ (\malogshort), \maloghard\ (\maloghardshort), which were derived by \citeauthor{MAFSB17} (\citeyear{MAFSB17}). In the following, these latter four domains are abbreviated to MBS. The difference w.r.t.\ the CoDMAP domains \blocksworld\ and \logistics\ is that for the domains of MBS many private actions need to be executed between two consecutive public actions, 
and agents must choose among several paths for achieving goals. All domains have uniform action costs. 

All tests are run on an InfiniBand Cluster with 512 nodes and 128 Gbytes of RAM, each node has two 8-cores Intel Xeon E5-2630 v3 with 2.40 GHz. Given a MA-planning problem, for each agent in the problem we limited the usage of resources to 3 CPU cores and 8 GB of RAM. Moreover, unless otherwise specified, the time limit was 5 minutes, after which the termination of all threads was forced.

 Table \ref{tab:coverage} shows the number of problems solved by \mabfws\ using seven different heuristics for the benchmark problems of CoDMAP and MBS. Planner \mabfws(\hff) is the baseline for our comparison, since it does not use novelty-based preferences to guide the search. For six out of sixteen domains, \mabfws\ with \hff\ solves almost all the problems. These are the domains with problems that require less interaction among agents. 

\mabfws\ with $f_1$ solves few more problems than \hff, and the domains where \mabfws$(f_1)$ performs well are the same as those with \hff. Surprisingly, \mabfws\ with $f_2$ solves many more problems than with \hff\ and $f_1$. The main difference between $f_2$ and $f_1$ is that the novelty-based exploration gives preference according to the number of unachieved goals $G_\bot$. This clearly results in a positive interplay with the search procedure. Interestingly, \mabfws\ with $f_2$ solves many problems of CoDMAP domains such as \logistics, \depot, and \woodworking, and several problems from benchmark MBS, which require a greater interaction among different agents.

\mabfws\ with $f_3$ or $f_4$ solves few more problems than with $f_2$, showing that the information about the number of unreachable goals from search states can be useful. 
Heuristic $f_5$ is computationally less expensive than $f_3$ and $f_4$, but the goal-directed component of $f_5$ is less accurate.
The results in Table \ref{tab:coverage} show that the tradeoff between computational cost and accuracy of $f_5$ does not pay off. \mabfws\ with $f_5$ is better than with \hffp\ and $f_1$, but solves fewer problems than with $f_2$, $f_3$, and $f_4$. The reason of this behavior is that, when the number of incoming messages is high, heuristic $f_5$ is computationally still quite expensive.

\begin{figure}[!t]
\includegraphics[scale=0.20]{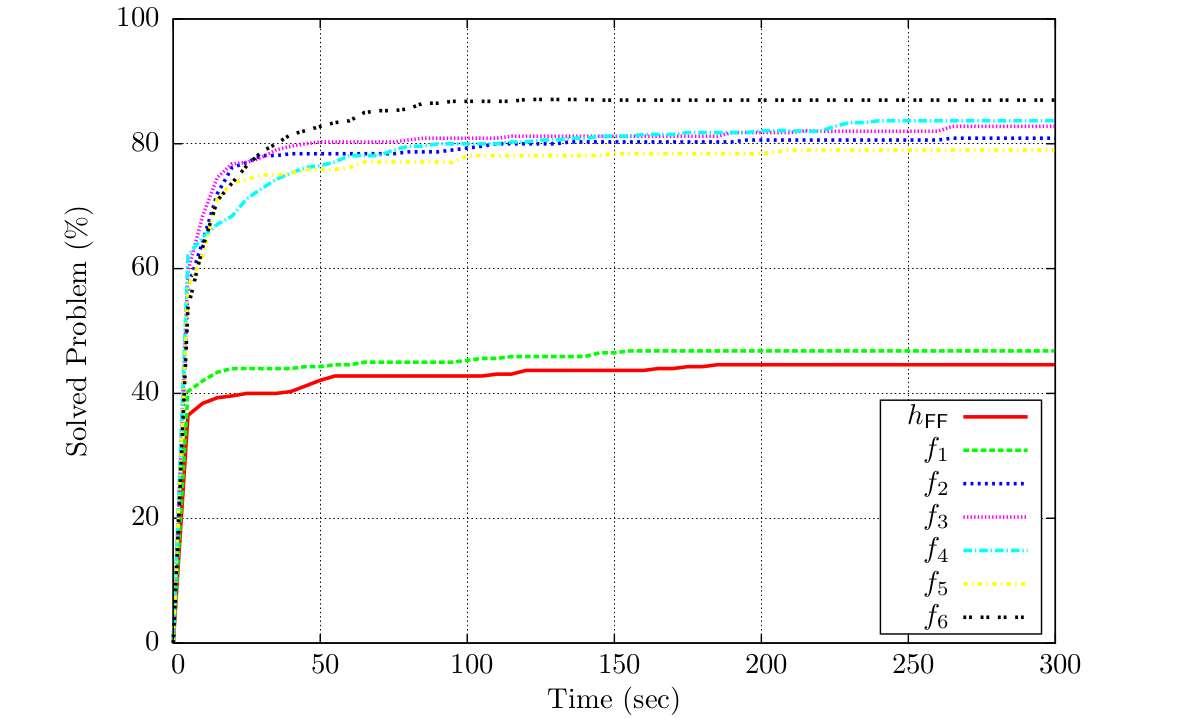}
\caption{Coverage as a function of the time for \mabfws\ using seven heuristics for benchmarks CoDMAP and MBS.}
\label{fig:coverage}
\end{figure}

The cheapest heuristic function to compute is $f_6$, since the most expensive step in the computation of our heuristics is the RPG construction and $f_6$ constructs the RPG only once. The results in Table \ref{tab:coverage} indicate that $f_6$ is a good tradeoff between accuracy and computational cost, since \mabfws\ with $f_6$ solves the largest set of problems. It solves several problems even for domain \mabw, which are unsolved by using any other heuristic function. 

Figure \ref{fig:coverage} shows the coverage of \mabfws\ with the seven heuristic functions using a time limit ranging from 0 to 300 seconds. With a time limit of few seconds, the best heuristic is $f_4$; with a time limit between 5 and 25 seconds, $f_3$ is the best; with a time limit between 25 seconds and 300 seconds \mabfws\ with heuristic $f_6$ solves the largest set of problems. Interestingly, the coverages obtained using a time limit of 150 seconds are substantially the same as 300 seconds. 

Table \ref{tab:metric} shows the performance of \mabfws\ using the proposed search heuristics in terms of average time, plan length, number of exchanged messages, number of expanded states, time score, and quality score. The averages are computed over the problems solved by all the compared heuristics. The time score and quality score are the measures originally proposed for the seventh international planning competition \cite{IPC7}. \mabfws\ with $f_4$ is on average the fastest, and the average numbers of exchanged message and expanded states of $f_4$ are therefore the lowest, followed closely by $f_3$. Remarkably, the average number of exchanged messages and expanded states of \mabfws\ with \hff\ and $f_1$ are almost two orders of magnitude greater than with the other heuristics. 

\begin{table}[!t]
\scriptsize
\begin{center}
\begin{tabular}{|l|c|c|c|c|c|c|c|}
\hline
Metric & \hff&
$f_1$&
$f_2$ &
$f_{3}$&
$f_{4}$&
$f_{5}$&
$f_{6}$ \\ \hline 
Avg.T&8.62&6.36&1.69&1.57&{\bf 1.55}&2.14&3.51 \\ 
Avg.L&61.4&{\bf 55.9}&61.4&60.3&61.0&92.8&67.0 \\ 
kMess&1749.7&1433.8&25.9&25.0&{\bf 24.8}&46.8&97.4 \\ 
kState&952.8&779.6&17.9&16.7&{\bf 16.2}&34.3&82.1 \\ \hline
Score\, Q&122.6&134.4&211.1&217.5&213.1&176.1&{\bf 220.8} \\ 
Score\, T&121.9&124.9&223.9&231.7&228.1&220.0&{\bf 233.4} \\ 
\hline 
\end{tabular}
\end{center}
\caption{\label{tab:metric}Average time, average plan length, number of exchanged messages (in thousands), number of expanded states (in thousands), time and quality score of \mabfws\ with seven heuristics for benchmarks CoDMAP and MBS.  
}
\end{table}

The limits of our approach are inherited from the width based algorithms. Width based algorithms such as \iw\ perform poorly in problems with high width. Variants such as \siw\ and \bfws\ try to mitigate the high width of the problems by using serialization or heuristics. When the novelty is used for pruning, the algorithms may become incomplete; if novelty is used as a preference, then completeness is not compromised. In general, novelty will help if the paths to the goal have low width, while problems that require reaching states with high width will become more challenging.

\begin{table}
\centering
\scriptsize
\begin{tabular}{|l|c|c|c|c|}
\hline
Domain & \masiw & \mabfws$(f_6)$ & \maplan & ~~\psm~~ \\
\hline
	\blocksworld &{\bf 20} &{\bf 20} &{\bf 20} &{\bf 20} \\ 
	\depot       &8        &{\bf 20} &12       & 17\\ 
	\driverlog   &{\bf 20} &{\bf 20} &16       &{\bf 20} \\ 
	\elevators   &{\bf 20} &{\bf 20} &8        & 12\\ 
	\logistics   &18       &{\bf 20} &18       & 18\\ 
	\rovers      &{\bf 20} &{\bf 20} &{\bf 20} & 19 \\ 
	\satellites  &{\bf 20} &{\bf 20} &{\bf 20} & 13\\ 
	\sokoban     &4        &{\bf 17} &{\bf 17} & 16 \\ 
	\taxi        &{\bf 20} &{\bf 20} &{\bf 20} &{\bf 20}\\ 
	\wireless    &0        &2        &{\bf 4}  & 0 \\ 
	\woodworking &1        &14       &{ 15}    &{\bf 18} \\ 
	\zenotravel  &{\bf 20} &{\bf 20} &{\bf 20} & 10 \\ 
	\hline 
{\bf Overall} (240)&171   &213      &190      & 184 \\ 
	\hline 
\end{tabular}
\caption{\label{tab:StateOfTheArt}Number of problems solved by  \masiw, \mabfws\ with heuristic $f_6$, \maplan, and \psm\ for benchmark CoDMAP.  The best performance is in bold. 
}
\end{table}

Finally, we compared our approach with other three existing approaches, \masiw, which is the approach mostly related to our work, the best configuration of \maplan, and \psm. \maplan\ and \psm\ are the best two planners that took part in the CoDMAP competition \cite{vstolba2015competition}. 
Table \ref{tab:StateOfTheArt} shows the results of this comparison for the CoDMAP domains. As for benchmark MBS, \masiw\ solves no problem, while \maplan\ and \psm\ do not support private goals, which are present in these problems. The time limit used for this comparison is 30 minutes, that is the same limit used in the competition. 

The results in Table \ref{tab:StateOfTheArt} show that for the competition problems \mabfws\ outperforms \masiw\ and is better than \maplan\ and \psm. 
Another planner we experimented is \dpp\ \cite{maliah2016stronger} that, to the best of our knowledge, is the state-of-the-art for the CoDMAP problems. We observed that, with our test environment, \mabfws$(f_6)$ solves few more problems than \dpp. 
Remarkably, the only type of information that the agents share by using our approach is the exchanged search states, while \maplan, \psm, and \dpp\ require sharing also the information for the computation of the search heuristics. In this sense, besides solving more problems, \mabfws\ preserves the agents' privacy more strongly than the other planners.

\section{Conclusions}

Goal-directed search is the main computational approach that has been investigated in classical planning and, subsequently, in MA planning. For classical planning, width-based exploration in the form of novelty-based preferences provides an effective complement to goal-directed search. 

In our setting for MA planning, in order to preserve privacy, we do not transmit the public projection of the actions, and hence the proposed goal-directed heuristics are not as informed as in classical planning. Moreover, the encryption of the private knowledge that the agents share during the search affects the measure of novelty.  Nevertheless, this work shows that the combination of new goal-directed heuristics computed efficiently and width-based search is also effective for MA planning. This opens up the possibility to increase privacy preserving properties of MA planning algorithms. For instance, given the success on black-box planning for single agents~\cite{frances2017purely}, we plan  to investigate the implications of fully protected models given as black-boxes, and the effect of novelty pruning in terms of sent messages and privacy.

\section*{Acknowledgements}

This research carried out with the support of resources of the National Collaborative Research Infrastructure Strategy (NeCTAR), and the Big \& Open Data Innovation Laboratory (BODaI-Lab) of the University of Brescia, which is granted by Fondazione Cariplo and Regione Lombardia. Nir Lipovetzky, has been partially funded by DST group.
\bibliographystyle{aaai}
\bibliography{biblio}

\end{document}